\documentclass[pdflatex,sn-mathphys-num]{sn-jnl}

\usepackage{graphicx}%
\usepackage{multirow}%
\usepackage{amsmath,amssymb,amsfonts}%
\usepackage{amsthm}%
\usepackage{mathrsfs}%
\usepackage[title]{appendix}%
\usepackage{xcolor}%
\usepackage{textcomp}%
\usepackage{manyfoot}%
\usepackage{booktabs}%
\usepackage{algorithm}%
\usepackage{algorithmicx}%
\usepackage{algpseudocode}%
\usepackage{listings}%
\usepackage{CJKutf8}
\usepackage{makecell}  

\theoremstyle{thmstyleone}%
\theoremstyle{thmstyletwo}%

\theoremstyle{thmstylethree}%

\begin{document}

\title[Article Title]{Can Large Language Models Capture Human Risk Preferences? A Cross-Cultural Study}


\author[1]{\fnm{Bing} \sur{Song}}\email{bing.song@connect.ust.hk}
\equalcont{These authors contributed equally to this work.}

\author[1]{\fnm{Jianing} \sur{Liu}}\email{jliudc@connect.ust.hk}
\equalcont{These authors contributed equally to this work.}

\author*[1]{\fnm{Sisi} \sur{Jian}}\email{cesjian@ust.hk}

\author*[2,3,4]{\fnm{Chenyang} \sur{Wu}}\email{cywu@nwpu.edu.cn}

\author[5]{\fnm{Vinayak} \sur{Dixit}}\email{v.dixit@unsw.edu.au}

\affil*[1]{\orgdiv{Department of Civil and Environmental Engineering}, \orgname{The Hong Kong University of Science and Technology}, \orgaddress{\street{Clear Water Bay, Kowloon}, \city{Hong Kong}, \postcode{999077},  \country{China}}}

\affil[2]{\orgdiv{School of Aeronautics}, \orgname{Northwestern Polytechnical University}, \orgaddress{\country{China}}}

\affil[3]{\orgdiv{Urban System Laboratory}, \orgname{Imperial College London}, \orgaddress{\country{United Kingdom}}}

\affil[4]{\orgdiv{National Key Laboratory of Aircraft Configuration Design}, \orgaddress{\country{China}}}

\affil[5]{\orgdiv{School of Civil and Environmental Engineering}, \orgname{The University of New South Wales}, \orgaddress{\country{Australia}}}


\abstract{Large language models (LLMs) have made significant strides, extending their applications to dialogue systems, automated content creation, and domain-specific advisory tasks. However, as their use grows, concerns have emerged regarding their reliability in simulating complex decision-making behavior, such as risky decision-making, where a single choice can lead to multiple outcomes. This study investigates the ability of LLMs to simulate risky decision-making scenarios. We compare model-generated decisions with actual human responses in a series of lottery-based tasks, using transportation stated preference survey data from participants in Sydney, Dhaka, Hong Kong, and Nanjing. Demographic inputs were provided to two LLMs—ChatGPT 4o and ChatGPT o1-mini—which were tasked with predicting individual choices. Risk preferences were analyzed using the Constant Relative Risk Aversion (CRRA) framework. Results show that both models exhibit more risk-averse behavior than human participants, with o1-mini aligning more closely with observed human decisions. Further analysis of multilingual data from Nanjing and Hong Kong indicates that model predictions in Chinese deviate more from actual responses compared to English, suggesting that prompt language may influence simulation performance. These findings highlight both the promise and the current limitations of LLM in replicating human-like risk behavior, particularly in linguistic and cultural settings.}

\keywords{Large language model, Risk attitude, Stated Preference Survey  }



\maketitle

\section{Introduction}\label{sec1}

Large language models (LLMs) have witnessed remarkable development in recent years \citep{nie2025joint}. Compared to earlier natural language processing tools, their capabilities have expanded far beyond traditional language processing tasks, extending to areas such as dialogue systems \citep{yi2024survey}, automated content creation \citep{ohde2025burden}, and specialized domain applications, including legal, financial, and medical advisory \citep{cheong2024not,challagundla2024financial,chen2024effect}. As a result, the trust placed in the outputs of LLMs is under increasing critical scrutiny, particularly regarding the underlying logic of the models and the reliability of the generated results.

As scrutiny of LLM outputs grows, researchers have begun to explore how LLMs perform in simulating human behavior, particularly in survey-based contexts \citep{liu2024can,xu2025morality}. In addition to traditional survey-style question-and-answer formats, role-playing language agents (RPLA) have emerged as a promising approach. These agents prompt LLMs to embody specific social roles—such as a doctor, policymaker, or consumer—and respond from the perspective of that role within a given scenario. This method allows for more context-sensitive and socially grounded behavior simulations, enhancing the ecological validity of experiments involving human-like decision-making. Existing studies have examined various human-like tasks, including value judgments \citep{tjuatja2024llms}, perceptual analysis \citep{li2024frontiers}, and intertemporal choices \citep{goli2024frontiers}, providing insights into the extent to which LLM responses align with human preferences. However, research to date has concentrated on relatively straightforward tasks, offering limited insight into how LLMs handle complex decision-making scenarios, such as choice situations involving risk and uncertainty.

Choice under risk and uncertainty refers to situations where respondents need to choose among multiple alternatives, each associated with different possible outcomes. Unlike straightforward factual queries, risk-based decisions require a comprehensive evaluation of all potential consequences linked to each option, making them a considerably more complex cognitive process \citep{kahneman2013prospect}. Such decisions are common in real-world contexts, influencing outcomes that range from individual financial choices \citep{eeckhoudt2005economic, engelmann2009expert} to government policymaking \citep{lafont2015deliberation, greenspan2004risk}. Consequently, simulating human behavior in risky scenarios allows us to assess how well LLMs replicate underlying decision-making processes. These simulations can, in turn, reveal potential limitations of LLMs—such as misjudging risk or exhibiting overly cautious behavior—which help define the boundaries of LLM capabilities.

To better understand how LLMs handle uncertainty, this study assesses the extent to which LLMs can replicate human behavior when faced with risky choices. Specifically, we compare the simulated choices made by LLMs to actual human decisions across multiple series of lottery choice games. Survey data were collected from participants in Sydney (Australia), Dhaka (Bangladesh), Hong Kong (China), and Nanjing (China), providing a cross-cultural basis for analysis. Demographic information—including age, gender, education, and income—was input into two LLMs (ChatGPT 4o and ChatGPT o1-mini), which were tasked with predicting the choices of corresponding participants. Using the Constant Relative Risk Aversion (CRRA) framework to calibrate risk preferences, we compare the risk preference estimated from the simulated behaviors with the actual responses of human participants. 

Our findings indicate that both LLMs exhibit more risk-averse behavior than human participants, with o1-mini aligning more closely with real-world observations than 4o. Additionally, for the data collected in Nanjing and Hong Kong, where multiple languages were used in the survey, we further examine the impact of languages on the simulated risk preferences \citep{boroditsky2001does, goli2024frontiers}. Our results show that when the models operate in Chinese, both LLMs deviate more significantly from reality compared to their English-based outputs, even though Chinese more closely aligns with the daily language used by respondents in these two datasets. Notably, o1-mini demonstrates greater robustness than 4o.

The remainder of this paper is organized as follows: Section 2 provides a review of the relevant literature. Section 3 outlines the dataset employed in this study. Section 4 presents the modeling framework. Section 5 discusses the empirical findings and their policy implications. Section 6 concludes the paper with key findings and directions for future research.

\section{Literature Review}

The rapid advancement of large language models (LLMs) has significantly accelerated the development of Role-Playing Language Agents (RPLAs) \citep{achiam2023gpt, anil2023gemini}. RPLAs enable artificial intelligence to interact with humans in ways that resemble embodied intelligence \citep{chen2024effect}, making these agents increasingly capable of simulating complex social behaviors. Through alignment training, RPLAs can replicate the knowledge systems of specific individuals, imitate their linguistic styles and behavioral tendencies, and reproduce latent personal characteristics \citep{dai2024mmrole, ge2024scaling, dai2023uncovering}. When combined with contextual prompting techniques, RPLAs can be tailored to simulate specific personas or emulate social groups by drawing on internal parameterized knowledge. 

These capabilities have made RPLAs valuable tools in scientific research, particularly in the simulation and modeling of human behavior. Existing studies in this area can be broadly categorized into two directions: (1) leveraging state-of-the-art LLMs to simulate scenarios that are difficult or infeasible to study in the real world; and (2) using RPLAs powered by different LLMs to perform established tasks and comparing their outputs with real-world human data to evaluate model alignment and performance.

Studies in the first category primarily focus on economic and healthcare contexts, where real-world observations are often hindered—by long time horizons and the inability to isolate causal mechanisms in the former, and by ethical regulations and privacy concerns in the latter. In the economics domain, Zhao et al. (2023) proposed CompeteAI, a GPT-4-powered framework simulating a virtual town with customer and restaurant agents, enabling the study of competitive dynamics that are otherwise difficult to capture due to the lack of granular behavioral data and the challenges of conducting controlled experiments in open markets \citep{zhao2023competeai}. Similarly, Li et al. (2024) developed EconAgent, a macroeconomic simulation framework in which RPLAs act as heterogeneous agents making work and consumption decisions. In the healthcare domain, Schmidgall et al. (2024) introduced AgentClinic, a simulation framework in which LLM-powered agents represent doctors, nurses, and patients engaged in collaborative treatment processes \citep{schmidgall2024agentclinic}. This work complements Agent Hospital by Li et al. (2024), which also models hospital-based human–agent interactions under controlled yet lifelike conditions \citep{li2024quantifying}.

The second line of research leverages RPLAs to evaluate the behavioral fidelity of different LLMs by comparing their simulated outputs with empirical human data on well-established tasks. These studies commonly follow a simulate–compare–validate approach, using RPLAs as experimental agents to evaluate model alignment under replicable conditions. This approach not only assesses alignment between model-generated and human responses, but also helps identify which LLMs are better suited for specific decision-making or perceptual tasks. For instance, \cite{xie2024travelplanner} compared the performance of various LLMs in travel planning scenarios using public datasets that included user-defined preferences, constraints, and itinerary goals, thereby evaluating each model’s ability to handle multi-objective decision-making under realistic conditions. \cite{li2024frontiers} examined how different LLMs simulated brand perception by prompting agents to evaluate well-known brands across dimensions such as trustworthiness and innovation, and quantitatively compared the generated outputs with large-scale consumer survey data to assess alignment with actual human attitudes. \cite{goli2024frontiers} assessed the performance of GPT-3.5 and GPT-4 in simulating human preference elicitation in structured survey tasks, using conjoint-style experiments and benchmarked model outputs against real-world human response datasets to evaluate how well model-driven choices mirrored actual human trade-offs. Comparable methods have also been applied in other fields, including theory of mind \citep{xu2024opentom}, recommendation systems \citep{lin2024data}, and code generation and sensory modeling \citep{coignion2024performance, mucha2024text2taste}.

\textbf{Research Gaps:} Despite the growing body of research demonstrating the utility of LLM-based RPLAs in behavioral science, the current literature still exhibits several notable limitations. First, the tasks explored in both lines of research—either simulating multi-agent interactions or benchmarking LLM behavior against human data—tend to focus on foundational problems with low complexity. Specifically, RPLAs are typically tasked with solving well-defined problems with clear objectives and minimal ambiguity. Consequently, the understanding of how RPLAs perform in more complex environments—particularly those involving risk, uncertainty, or dynamic trade-offs—remains limited. 

Second, the empirical data used for evaluation is often drawn from narrow sources, typically representing a single cultural or linguistic context. Systematic investigation into how LLM-based agents behave across different populations, countries, or languages when faced with similar tasks are still lacking. This absence of cross-cultural and cross-linguistic comparisons leaves important questions unanswered about the generalizability and robustness of these models in diverse real-world settings.

\section{Data}
This study utilized four datasets collected through surveys conducted with real respondents in Sydney, Hong Kong, Dhaka, and Nanjing. These four datasets originate, respectively, from the following studies: \cite{dixit2019risk}, \cite{liu2024risky}, \cite{dixit2019eliciting}, and \cite{Guo2025}. All four contain questions on respondents' socio-demographic features and include lottery-choice games designed to elicit respondents' risk attitudes. While each dataset contains rich information about respondents, the specific attributes available vary slightly due to differences in survey design and local priorities. For example, some include detailed employment or household composition variables, while others do not. To ensure consistency and comparability in simulated agents, this study uses the set of socio‑demographic variables common across all four datasets: age, gender, education level and income. 

\subsection{Descriptive Analysis of Socio-demographics}
A descriptive analysis of the socio‑demographic variables shared across the four datasets is presented in Table \ref{tab:Descriptive_Stats}. Substantial differences are observed across the four datasets: on average, the Sydney sample is the youngest, while the Hong Kong sample is the oldest. The Nanjing dataset exhibits the highest average education level. Notably, the Dhaka sample is distinct due to its low variability across all four variables, reflecting the sample’s primarily focus on taxi drivers during data collection.

\begin{table}[h]
    \caption{Descriptive Statistics of Socio-Demographic Variables Used in This Study Across the Four Datasets}
    \label{tab:Descriptive_Stats}
    \begin{tabular}{lllll}
        \toprule
        & Age (SD) & Male (\%) & Bachelor or Above (\%) & Income (SD) \\
        \midrule
        Sydney     & 28.7 (25.5)  & 79.70\%  & 50\%    & 3.8 (2.6) \\
        Hong Kong  & 40.7 (13.1)  & 45.30\%  & 62.50\% & 3.8 (1.4) \\
        Dhaka      & 37.4 (7.5)   & 100\%    & 0\%     & 1.9 (0.3) \\
        Nanjing    & 36.8 (9.2)   & 50.30\%  & 85.50\% & 2.5 (1.0) \\
        \bottomrule
    \end{tabular}
    \par\vspace{2mm}
    \begin{minipage}{\linewidth}
        \footnotesize \textit{Note}: Due to differences in exchange rates and local purchasing power, the ``income'' variable is not directly comparable across the four datasets. Instead, the values in this column should be interpreted as indicative of the relative relationship between the mean and standard deviation within each region.
    \end{minipage}
\end{table}

\subsection{Overview of Lottery Choice Games}
A lottery game is a controlled experimental task in which participants choose between probabilistic outcomes. By systematically varying the probabilities and magnitudes of these outcomes, each individual’s risk attitude—that is, their degree of aversion to or preference for risk—can be inferred. The setup of the lottery choice game in each survey is detailed below. It is worth noting that while the outcome values of the lotteries vary across the four surveys, the primary focus should be on the differences in expected values between the left and right options within each game.

The Sydney dataset consists of a series of nine lottery choice tasks. Each task presents participants with a binary decision: the right option offers a fixed outcome, while the left option involves risk, providing either a higher or lower reward. The probabilities associated with the left options are clearly communicated to participants and vary across tasks. Details of these lottery tasks are presented in Table \ref{tab:Sydney dataset}. The terms $EV_{\text{Left}}$ and $EV_{\text{Right}}$ represent the expected utility of selecting the left and right lotteries, respectively. The Sydney dataset includes responses from 64 valid participants.

\begin{table}[h]
    \caption{Lottery Setup for the Sydney Dataset.}\label{tab:Sydney dataset}
    \begin{tabular}{lllllllll}
        \toprule
        Lottery & \multicolumn{3}{l}{Left Lottery} & \multicolumn{3}{l}{Right Lottery} & $EV_{\text{Left}}$ & $EV_{\text{Right}}$ \\
        \cmidrule(lr){2-4} \cmidrule(lr){5-7}
         & Low & High & $P_{\text{Low}}$ & Fixed &  &  & & \\
        \midrule
        1 & 0.5 & 20 & 0.5 & 8 &  &  & 10.25 & 8 \\
        2 & 0.5 & 20 & 0.7 & 1 &  &  & 6.35 & 1 \\
        3 & 0.5 & 20 & 0.9 & 4 &  &  & 2.45 & 4 \\
        4 & 0.5 & 20 & 0.5 & 6 &  &  & 10.25 & 6 \\
        5 & 0.5 & 20 & 0.7 & 2 &  &  & 6.35 & 2 \\
        6 & 0.5 & 20 & 0.9 & 2 &  &  & 2.45 & 2 \\
        7 & 0.5 & 20 & 0.5 & 4 &  &  & 10.25 & 4 \\
        8 & 0.5 & 20 & 0.7 & 5 &  &  & 6.35 & 5 \\
        9 & 0.5 & 20 & 0.9 & 1 &  &  & 2.45 & 1 \\
        \bottomrule
    \end{tabular}
\end{table}

The Hong Kong dataset also consists of nine lottery choice tasks. The probability settings are identical to those in \cite{dixit2019risk}; however, the lottery values have been scaled by a factor of 100, as detailed in Table \ref{tab:HK dataset}. This dataset includes responses from 997 valid participants.

\begin{table}[h]
    \caption{Lottery Setup for the Hong Kong Dataset.}\label{tab:HK dataset}
    \begin{tabular}{lllllllll}
        \toprule
        Lottery & \multicolumn{3}{l}{Left Lottery} & \multicolumn{3}{l}{Right Lottery} & $EV_{\text{Left}}$ & $EV_{\text{Right}}$ \\
        \cmidrule(lr){2-4} \cmidrule(lr){5-7}
         & Low & High & $P_{\text{Low}}$ & Fixed &  &  & & \\
        \midrule
        1 & 50 & 2000 & 0.5 & 800 &  &  & 1025 & 800 \\
        2 & 50 & 2000 & 0.7 & 100 &  &  & 635 & 100 \\
        3 & 50 & 2000 & 0.9 & 400 &  &  & 245 & 400 \\
        4 & 50 & 2000 & 0.5 & 600 &  &  & 1025 & 600 \\
        5 & 50 & 2000 & 0.7 & 200 &  &  & 635 & 200 \\
        6 & 50 & 2000 & 0.9 & 200 &  &  & 245 & 200 \\
        7 & 50 & 2000 & 0.5 & 400 &  &  & 1025 & 400 \\
        8 & 50 & 2000 & 0.7 & 500 &  &  & 635 & 500 \\
        9 & 50 & 2000 & 0.9 & 100 &  &  & 245 & 100 \\
        \bottomrule
    \end{tabular}
\end{table}

The Dhaka dataset comprises ten lottery choice tasks. Each task presents participants with a binary decision, where both options involve risk. In the left lottery, participants could receive a lower outcome of 6 with probability \( p \), or a higher outcome of 8 with probability \( 1-p \). Similarly, the right lottery offers a lower outcome of 1 with probability \( p \), or a higher outcome of 20 with probability \( 1-p \). The value of \( p \) varies across tasks, with details provided in Table \ref{tab:Dhaka dataset}, where \( p \) corresponds to the “Prob. payoff 2” column. The Dhaka dataset includes responses from 101 valid participants.

\begin{table}[h]
    \caption{Lottery Setup for the Dhaka Dataset.}\label{tab:Dhaka dataset}
    \begin{tabular}{llllllllll}
        \toprule
        Lottery & \multicolumn{2}{l}{Prob. payoff} & \multicolumn{2}{l}{Left Lottery} & \multicolumn{2}{l}{Right Lottery} & $EV_{\text{Left}}$ & $EV_{\text{Right}}$ \\
        \cmidrule(lr){2-3} \cmidrule(lr){4-5} \cmidrule(lr){6-7}
        & 1 & 2 & Payoff 1 & Payoff 2 & Payoff 1 & Payoff 2 & & \\
        \midrule
        1  & 0.1 & 0.9 & 8 & 6 & 20 & 1 & 6.2 & 2.9 \\
        2  & 0.2 & 0.8 & 8 & 6 & 20 & 1 & 6.4 & 4.8 \\
        3  & 0.3 & 0.7 & 8 & 6 & 20 & 1 & 6.6 & 6.7 \\
        4  & 0.4 & 0.6 & 8 & 6 & 20 & 1 & 6.8 & 8.6 \\
        5  & 0.5 & 0.5 & 8 & 6 & 20 & 1 & 7.0 & 10.5 \\
        6  & 0.6 & 0.4 & 8 & 6 & 20 & 1 & 7.2 & 12.4 \\
        7  & 0.7 & 0.3 & 8 & 6 & 20 & 1 & 7.4 & 14.3 \\
        8  & 0.8 & 0.2 & 8 & 6 & 20 & 1 & 7.6 & 16.2 \\
        9  & 0.9 & 0.1 & 8 & 6 & 20 & 1 & 7.8 & 18.1 \\
        10 & 1   & 0   & 8 & 6 & 20 & 1 & 8.0 & 20.0 \\
        \bottomrule
    \end{tabular}
\end{table}

The Nanjing dataset consists of ten lottery choice tasks, similar to those in the Dhaka dataset, and follows the same probability structure. However, the payoffs for the left and right lotteries differ, as detailed in Table \ref{tab:Nanjing dataset}. The Nanjing dataset includes responses from 145 valid participants.

\begin{table}[h]
    \caption{Lottery Setup for the Nanjing Dataset.}\label{tab:Nanjing dataset}
    \begin{tabular}{llllllllll}
        \toprule
        Lottery & \multicolumn{2}{l}{Prob. payoff} & \multicolumn{2}{l}{Left Lottery} & \multicolumn{2}{l}{Right Lottery} & $EV_{\text{Left}}$ & $EV_{\text{Right}}$ \\
        \cmidrule(lr){2-3} \cmidrule(lr){4-5} \cmidrule(lr){6-7}
        & 1 & 2 & Payoff 1 & Payoff 2 & Payoff 1 & Payoff 2 & & \\
        \midrule
        1  & 0.1 & 0.9 & 2 & 1.6 & 3.85 & 0.1 & 1.64 & 0.475 \\
        2  & 0.2 & 0.8 & 2 & 1.6 & 3.85 & 0.1 & 1.68 & 0.85 \\
        3  & 0.3 & 0.7 & 2 & 1.6 & 3.85 & 0.1 & 1.72 & 1.225 \\
        4  & 0.4 & 0.6 & 2 & 1.6 & 3.85 & 0.1 & 1.76 & 1.6 \\
        5  & 0.5 & 0.5 & 2 & 1.6 & 3.85 & 0.1 & 1.8 & 1.975 \\
        6  & 0.6 & 0.4 & 2 & 1.6 & 3.85 & 0.1 & 1.84 & 2.35 \\
        7  & 0.7 & 0.3 & 2 & 1.6 & 3.85 & 0.1 & 1.88 & 2.725 \\
        8  & 0.8 & 0.2 & 2 & 1.6 & 3.85 & 0.1 & 1.92 & 3.1 \\
        9  & 0.9 & 0.1 & 2 & 1.6 & 3.85 & 0.1 & 1.96 & 3.475 \\
        10 & 1   & 0   & 2 & 1.6 & 3.85 & 0.1 & 2 & 3.85 \\
        \bottomrule
    \end{tabular}
\end{table}

\section{Methodology}
\subsection{Role-Playing Language Agents Architecture and Construction (RPLA)}

This subsection introduces the construction and structure of the RPLA used in this study. The RPLA is designed to simulate human respondents by assigning different roles and background profiles to an LLM through carefully crafted prompts \citep{shanahan2023role}. The RPLA consists of four key components: profile, memory, planning, and action \citep{chen2024persona}, as detailed below.

\textbf{Profile:} A profile defines the unique characteristics of a simulated individual, serving as the foundation for modeling realistic agent behaviors in RPLA systems \citep{chen2024roleinteract}. It typically captures several elements, including the agent’s inherent attributes (such as age, gender, or socio-economic background), task-related specifications (such as role identity, goal orientation, or decision criteria), and external constraints (such as cultural norms, environmental factors, or task-specific limitations). A well-constructed profile enables the agent to behave in a consistent and plausible manner within the simulated environment \citep{takagi2025framework}. In this study, we utilized survey data from four cities to construct simulated profiles for RPLAs. We use the four attributes introduced in Section 3.1 (age, gender, education level, and personal income)  for profile construction. Additionally, we treated the city of residence as a crucial contextual attribute, as it reflects broader socio-cultural and geographical environments that influence agent behavior and language use. 

The concept of a profile involves two key dimensions: form and construction \citep{xu2024character}. The form dimension refers to the structure and presentation of profile information—ranging from structured data formats (e.g., JSON schemas or attribute tables) to natural language descriptions or interactive prompts. The construction dimension, on the other hand, concerns how the information in the profile is obtained. This may involve manual creation by researchers, automatic generation via statistical or machine learning models, or derivation from empirical sources such as ethnographies or behavioral datasets. In this study, the form dimension is implemented as interactive prompts that generate natural language profile descriptions by combining demographic and contextual information into coherent, human-readable inputs. The construction dimension is implemented by extracting, standardizing, and aligning selected attributes from the four datasets to build a unified input framework for simulation. A sample English-language prompt is provided below to illustrate the final profile form used in our study.

\textit{You will take on the role of a survey respondent. The purpose of this survey is to assess people's risk attitudes through nine consecutive lottery questions. For each question, you are required to carefully consider and compare the expected income with the potential risks; however, you do not need to disclose your reasoning; simply provide your answers. In this survey, you will portray a {age}-year-old \{gender\} from \{city\}, \{country\}, who has completed the highest level of education: \{education\}. Your monthly income is: \{income\}.}

In addition to generating profiles in English, we also created profiles in Chinese to better align with the native language of respondents from Hong Kong and Nanjing, China. Prior research suggests that language differences can influence both the reasoning processes of RPLAs and experimental outcomes \citep{goli2024frontiers}. Therefore, we aim to investigate whether the language used in profile construction affects the RPLA’s risk preferences.

The sample prompt in Chinese, provided below, was translated from the English version.

\begin{CJK}{UTF8}{gbsn} 
你将承担调查受访者的角色。本调查的目的是通过一系列9个连续的彩票问题来评估个人的风险态度。对于每个问题，你需要仔细考虑并比较预期收入与潜在风险；然而，你无需向我透露你的思考过程——只需提供你的答案。在本调查中，你将扮演一位来自\{国家\}\{城市\}的\{年龄\}的\{性别\}，已完成最高学历：\{教育程度\}。你的月收入为：\{收入信息\}。
\end{CJK}

\textbf{Memory:} The second key component is memory. A well-known limitation of LLMs is their restricted context window, which constrains the amount of prior information they can retain during interactions. To maintain the consistency and coherence of RPLA behavior, it is therefore essential to design a memory mechanism capable of storing both received input and generated output from the agent \citep{alizadeh2024llm}.

Memory involves two aspects: memory types and memory operations. Memory can be broadly divided into short-term memory and long-term memory, with the former focusing on immediate, session-specific information and the latter concerning persistent knowledge across sessions. Memory operations refer to the agents' continuous updating and using their memory, including writing, retrieval, and reflection.
In this study, we use memory writing to ensure that RPLA maintains a short-term memory which stores transient information received during the interaction, such as user instructions, dialogue history, and user feedback. This design ensures that the agent can consistently make decisions in the current round of the lottery scenario based on the input profile and its previous choices throughout the session. Long-term memory across sessions, however, is intentionally excluded. This approach aligns with the natural thought process of real human respondents in real-world scenarios, as they have memory of their previous options in the survey, but lack access to others' experiences, and are required to complete the survey independently.

\textbf{Planing:} Planning refers to the process by which an agent formulates a sequence of actions to achieve specific objectives. In this study, we adopt empathetic planning, which leverages chain-of-thought (CoT) reasoning to enable the agent to adjust its decisions based on character-related information \citep{mou2024individual}. This approach enhances the agent’s ability to anticipate and infer the behaviors and emotions of simulated human respondents before making decisions.

We aim for the RPLA to simulate the decision-making processes of respondents when faced with sequential lottery-related questions. Specifically, we expect the agent to construct its reasoning chain by integrating information from its individual profile, local average income, its financial status, and the expected returns of different lottery choices. This ensures that the RPLA’s responses align more closely with the cognitive processes observed in human participants in real-world settings. We implement this process by inputting relevant information into the prompt. An example of such a prompt is as follows:

\textit{You need to carefully consider the question based on your background, with a particular focus on comparing the differences in expected values between the options.}

\textbf{Action:} Action refers to the direct interaction between RPLA and the real world. As an interface for simulating human behavior, action enables the RPLA to effectively execute tasks and generate responses. We employ a closed-domain approach and regulate the RPLA’s actions through prompt-based instructions. Specifically, when responding to lottery-related questions, the RPLA is required to provide only its final choice without revealing the underlying reasoning process. This design serves two primary purposes. First, it minimizes the influence of linguistic variability and noise that often emerge during open-ended reasoning generation. Since large language models are highly sensitive to prompt phrasing and contextual cues, exposing step-by-step thought processes may introduce inconsistencies unrelated to the core decision logic \citep{xu2022exploring}. Second, by limiting the output to a discrete choice, we standardize the response format, which facilitates reliable and efficient modeling of risk preferences across different scenarios and agents.An example of an action-related prompt is provided below:

\textit{However, you do not need to explain your reasoning process—simply indicate whether you choose "Option 1" or "Option 2."}

\subsection{Risk Attitudes Estimation}
A decision-maker’s risk attitude refers to their consistent pattern of ranking and choosing among uncertain outcomes, reflecting their willingness to accept variability in results \citep{pratt1978risk}. Specifically, when presented with two alternatives of equal expected value (e.g., a guaranteed gain of 50 USD versus a 50\% chance of gaining 100 USD), an individual who prefers the former is considered risk averse (avoiding potential risks), while one who prefers the latter is risk seeking (pursuing higher potential returns despite the risks). If the individual is indifferent between the two, they are considered risk neutral. In the Expected Utility framework, risk attitudes are represented by the curvature of an individual’s utility function over wealth: concave utility indicates risk aversion, linear utility indicates risk neutrality, and convex utility indicates risk seeking.

In this study, we employ the Constant Relative Risk Aversion (CRRA) model to quantify respondents' risk preferences. The utility function in the CRRA model is defined as:
\begin{equation}
    U(x) = \frac{x^{1-r}}{1-r}
\end{equation}
where \(x\) represents the reward and \(r\) indicates the risk preference  (\(r > 0\): risk-averse; \(r = 0\): risk-neutral; \(r < 0\): risk-seeking).

Building upon the framework from \cite{liu2023understanding}, we applied the equation \(r = r_0 + \alpha_1 X_1\) to reflect the effect of socio-demographic variables on risk preferences. In this equation, \(r_0\) is a constant risk parameter, while \(\alpha_1\) is a vector of coefficients that correspond to the socio-demographic variables \(X_1\), which include age, gender, education, and income.

\section{Empirical Analysis}
This section first compares risk attitudes estimated from the real and simulated data, and then examines the impact of using different prompt languages. To mitigate the impact of randomness in LLM-generated outputs, each lottery choice task faced by each respondent in each dataset was simulated three times. The majority output—i.e., the option selected in at least two out of three simulations—was adopted as the final simulated response. 

\subsection{Comparison of Real and Simulated Risk Attitudes}
We conducted paired t-tests to assess whether there were significant differences between risk attitudes estimated from real data (hereafter referred to as "real risk attitudes") and those estimated from LLM-simulated data (hereafter referred to as "simulated risk attitudes"). Table \ref{tab:Results_real_4o} compares real risk attitudes with those derived from ChatGPT 4o-simulated data, while Table \ref{tab:Results_real_o1} presents a comparison between real risk attitudes and estimates from ChatGPT o1-mini-simulated data.  

Across all four datasets, the risk attitudes estimated from ChatGPT 4o-simulated data are consistently higher than their corresponding real risk attitudes, indicating that the model exhibits a greater tendency toward risk-averse choices compared to humans. According to Table \ref{tab:Results_real_o1}, while the risk attitudes estimated from ChatGPT o1-mini-simulated data also differ significantly from real risk attitudes, they are notably closer to real values than those estimated by ChatGPT 4o. Furthermore, as shown in Table \ref{tab:Results_4o_o1}, this difference between the two simulated risk attitudes is also statistically significant across all datasets except for Dhaka, where the two models yield statistically indistinguishable risk attitudes. We infer that the ChatGPT o1-mini model's lack of significantly improved performance on the Dhaka dataset is likely due to the unique characteristics of that dataset. While the other three datasets were collected from a broad sociodemographic group, the Dhaka data was originally gathered exclusively from taxi drivers. Consequently, the Dhaka dataset exhibits limited gender variation (all participants are male), and both income and education levels are more narrowly distributed compared to the other datasets. We hypothesize that this lack of diversity may have limited the predictive performance of the o1-mini model.

\begin{table}[h]
    \caption{Estimation of Risk Attitudes Based on Real and ChatGPT 4o-simulated Data}
    \label{tab:Results_real_4o}
    \centering
    \begin{tabular}{llll}
        \toprule
        & Real & ChatGPT 4o & Ha: mean(diff) != 0 \\
        \midrule
        Sydney    & 0.420 & 1.387 & ***  \\
        Hong Kong & 0.765 & 1.620 & *** \\
        Dhaka     & 0.310 & 0.825 & ***  \\
        Nanjing   & 0.130 & 0.357 & *** \\
        \bottomrule
    \end{tabular}
    \par\vspace{2mm}
    \begin{minipage}{\linewidth}
        \footnotesize \textit{Note}: The level of confidence on significance is indicated by *** (\(p < 0.01\)), ** (\(0.01 \leq p < 0.05\)), and * (\(0.05 \leq p < 0.1\)).
    \end{minipage}
\end{table}

\begin{table}[h]
    \caption{Estimation of Risk Attitudes Based on Real and ChatGPT o1-mini-simulated Data}
    \label{tab:Results_real_o1}
    \begin{tabular}{llll}
        \toprule
        & Real & ChatGPT o1-mini & Ha: mean(diff) != 0 \\
        \midrule
        Sydney    & 0.420 & 0.641 & ***  \\
        Hong Kong & 0.765 & 0.509 & *** \\
        Dhaka     & 0.310 & 0.802 & ***  \\
        Nanjing   & 0.130 & 0.229 & *** \\
        \bottomrule
    \end{tabular}
\end{table}

\begin{table}[h]
    \caption{Comparison of Simulated Risk Attitudes by ChatGPT 4o and o1-mini}
    \label{tab:Results_4o_o1}
    \begin{tabular}{llll}
        \toprule
        & ChatGPT 4o & ChatGPT o1-mini & Ha: mean(diff) != 0 \\
        \midrule
        Sydney    & 1.387 & 0.641 & ***  \\
        Hong Kong & 1.620 & 0.509 & *** \\
        Dhaka     & 0.825 & 0.802 & 0.311  \\
        Nanjing   & 0.357 & 0.229 & *** \\
        \bottomrule
    \end{tabular}
\end{table}

\begin{figure}[H]
	\centering
		\includegraphics[scale=.2]{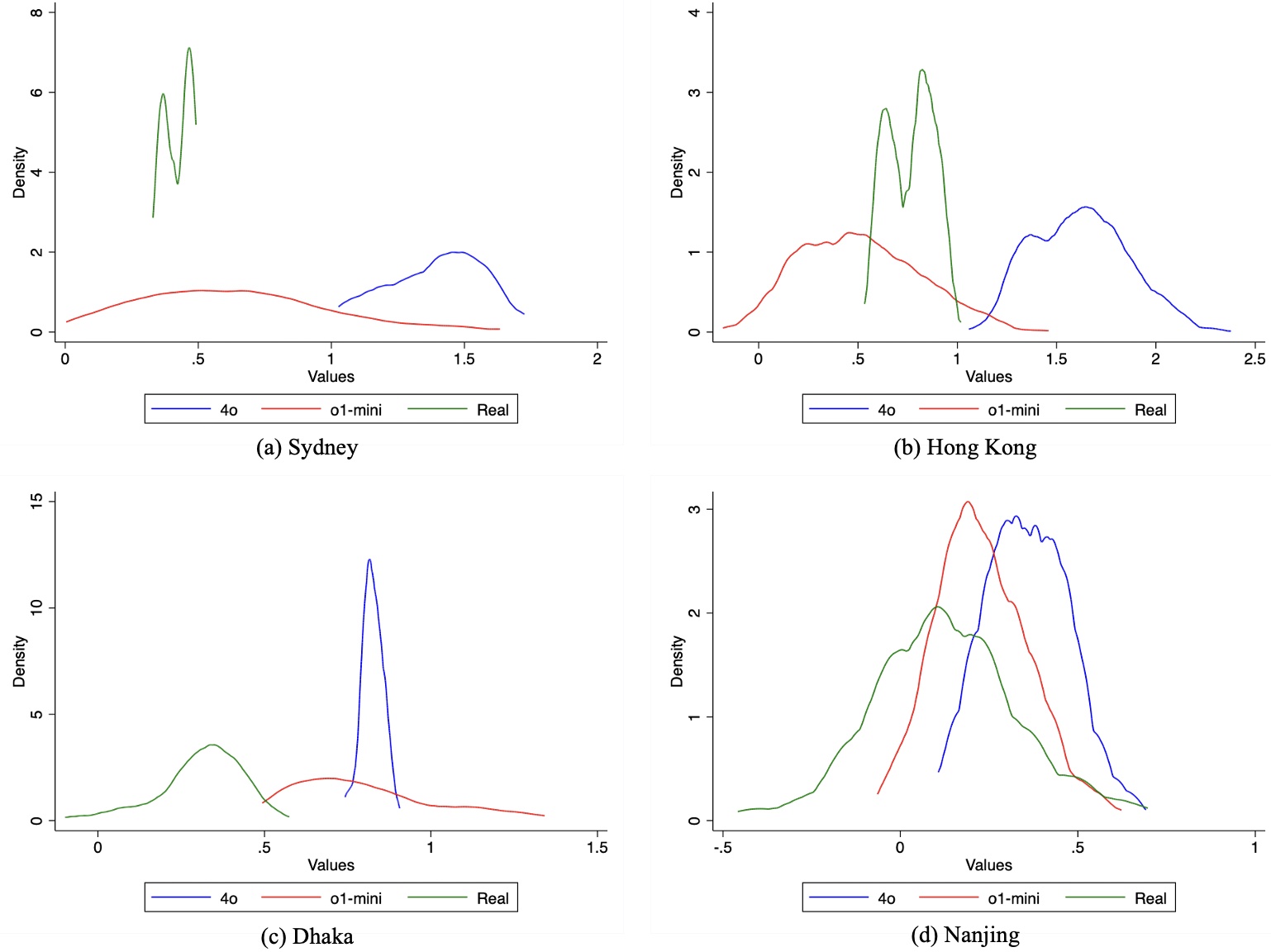}
	\caption{Kernel Density Plot of Risk Attitudes Estimated from Real and Simulated Data Across Locations: (a) Sydney, (b) Hong Kong, (c) Dhaka, and (d) Nanjing}
	\label{FIG:1}
\end{figure}

To further compare real and simulated risk attitudes, we analyzed the kernel density distribution of each dataset, as shown in Figure \ref{FIG:1}. The four subplots (a–d) correspond to the four cities, respectively.

In Sydney (subplot (a)), real risk attitudes are center around 0.42, while the distribution of ChatGPT 4o-simulated risk attitudes is noticeably shifted to the right, indicating a higher level of risk aversion than observed among real participants. Although the distribution of ChatGPT o1-mini-simulated risk attitudes still deviates from that of real participants, its average aligns more closely with the real distribution. In Hong Kong (subplot (b)), ChatGPT 4o-simulated risk attitudes again lie significantly to the right, reflecting substantially greater risk aversion compared to actual human data. In contrast, ChatGPT o1-mini-simulated risk attitudes exhibit a higher degree of overlap with the real distribution, though some deviation remains. In Dhaka (subplot c), both ChatGPT 4o- and o1-mini-simulated risk attitudes are positioned to the right of the human distribution, indicating a systematic overestimation of risk aversion. In Nanjing (subplot d), real participants exhibit a mean CRRA of approximately 0.13, suggesting near-risk-neutral or slightly risk-averse behavior. In comparison, both ChatGPT 4o- and o1-mini-simulated distributions once again shift to the right, demonstrating an overestimation of risk aversion. However, the o1-mini distribution consistently remains closer to the real data than that of 4o, reflecting the same trend observed across the other cities.

Across all four locations, the kernel density curves reveal consistent patterns in how LLM-simulated risk attitudes compare to real human data. In each city's plot, the green distribution (real participants) generally occupies a different position along the risk attitude axis than the blue (ChatGPT 4o) and red (ChatGPT o1-mini) curves. A clear trend emerges: LLM simulations exhibit systematic offsets, particularly with ChatGPT 4o.  Specifically, the LLM-generated distributions are consistently right-shifted relative to the real data, indicating that the LLMs tend to predict higher risk aversion than observed in real participants. This pattern appears in all four cities, suggesting a broad, systematic bias rather than a location-specific anomaly. 

Additionally, the shape of the LLM distributions appears similar across cities, typically resembling a normal distribution. The shape of the green curves (real data), however, vary more noticeably, sometimes even approaching a bimodal distribution. This suggests that LLMs apply a relatively generic, ``one-size-fits-all'' risk attitude, failing to account for cultural and contextual differences in human risk preferences.

From a policy perspective, this divergence raises concerns about directly applying LLM-based simulations to risk-sensitive policymaking. Without proper calibration, policymakers may overestimate the prevalence of risk-averse behavior in the population, potentially resulting in overly conservative policy measures that fail to accurately reflect actual behavior.

\subsection{Comparison of Simulated Risk Attitudes Across Different Language Prompts}
Given that the Nanjing and Hong Kong datasets were collected in non-English languages, we examined whether linguistic differences have impacts on simulated risk attitudes. Specifically, we explored the effect of using English and Chinese prompts on the simulated risk attitudes generated by ChatGPT 4o and ChatGPT o1-mini for these datasets. The results are presented in Table \ref{tab:Results_CH_EN}.


\begin{table}[h]
    \caption{Estimation of Risk Attitudes Using Different Language Prompts}
    \label{tab:Results_CH_EN}
    \centering
    \footnotesize  
    \begin{tabular}{llllllll}
        \toprule
        & Real & \multicolumn{3}{c}{ChatGPT 4o} & \multicolumn{3}{c}{ChatGPT o1-mini} \\
        \cmidrule(lr){3-5} \cmidrule(lr){6-8}
        & & English & Chinese & \makecell{Ha:\\mean(diff) $\neq$ 0} & English & Chinese & \makecell{Ha:\\mean(diff) $\neq$ 0} \\
        \midrule
        Hong Kong & 0.765 & 1.620 & 1.654 & *** & 0.509 & 0.390 & *** \\
        Nanjing   & 0.130 & 0.357 & 0.951 & *** & 0.229 & 0.250 & *** \\
        \bottomrule
    \end{tabular}
    \normalsize  
\end{table}

The results highlight that linguistic framing (English vs. Chinese) significantly influences LLM-simulated risk attitudes. Across all tested cases, using Chinese prompt led to greater deviation from real risk attitudes compared to using English prompt. This effect is evident in that, regardless of whether English prompts tend to overestimate or underestimate respondents' risk aversion, switching to Chinese prompt consistently amplifies the bias. Notably, despite most original data being collected in Chinese, using Chinese prompt does not improve alignment between simulations and real-world data. 

One possible explanation for this discrepancy is that LLMs do not fully capture the linguistic structures and cultural cues embedded in Chinese. Their predominantly English-based training may lead to an incomplete contextual understanding, thereby amplifying biases rather than approximating empirically observed risk attitudes. This highlights the importance of evaluating LLM outputs in cross-linguistic contexts: even if the original data were collected in Chinese, a model inadequately trained in the language may misrepresent real-world decision-making patterns.

From a policy perspective, these findings highlight the urgent need to develop and calibrate language-specific models that better reflect local cultural and linguistic structures. Policymakers and researchers should exercise caution when using LLM-generated outputs for risk attitude assessment, particularly in multilingual environments. It is essential to invest in enhancing model training by incorporating more diverse datasets and richer contextual cues from non-English languages. 

\section{Discussion and Conclusion}

Although LLMs have made significant advancements in recent years, their ability to simulate complex decision-making behavior, such as risky decision-making, remains unvalidated. This study explored the capacity of LLMs, specifically ChatGPT 4o and ChatGPT o1-mini, to replicate human risk-related decision-making behavior under uncertainty, using lottery-choice scenarios across four geographically and socioeconomically distinct datasets. Respondents’ risk attitudes were estimated using the CRRA model, based on both real survey data and simulated responses generated by the LLMs.

\subsection{Key Results}

First, both ChatGPT 4o and ChatGPT o1-mini systematically exhibited higher levels of risk aversion compared to actual human respondents in general, suggesting an intrinsic bias toward conservative decision-making. ChatGPT 4o consistently showed a stronger risk-averse tendency than ChatGPT o1-mini, indicating model-specific disparities in simulating human risk preferences. Notably, ChatGPT o1-mini generated estimates that were significantly closer to real human data.

Second, the results underscored the importance of sociodemographic variability. The limited demographic diversity within the Dhaka dataset appeared to constrain the predictive performance of ChatGPT o1-mini, resulting in risk attitude estimates that were indistinguishable from those produced by ChatGPT 4o. This finding suggests that LLMs may have been trained on relatively generic resources in terms of risk attitudes, and therefore their capabilities to simulate human risk preference while accounting for cultural and contextual differences in human risk preferences is weak.

Third, linguistic framing—specifically the use of English versus Chinese—significantly influenced the simulated risk attitudes. Counterintuitively, despite Chinese being inherently closer to the linguistic context of the respondents, using Chinese prompts led to greater deviations from the real data. This suggests that the LLMs tested, which are predominantly trained on English-language corpora, may fail to adequately capture the subtle linguistic and cultural nuances embedded in non-English languages.

\subsection{Implications}

The systematic biases observed in LLM-generated risk attitudes highlight a major limitation of relying on generalized LLMs for behavioral prediction in policy-relevant contexts. The consistent trend toward higher risk aversion may cause policymakers to overestimate conservatism in public behavior, potentially leading to overly cautious policies that do not align with actual societal risk-taking tendencies. It is therefore essential that any policy formulation based on LLM-generated data be verified via rigorous validation procedures, ideally involving comparisons with localized empirical data.

Furthermore, the findings from the Dhaka dataset reinforce the necessity of accounting for sociodemographic heterogeneity when using AI simulations for policy analysis. LLMs appear less capable of capturing nuanced risk preferences within demographically homogeneous groups, which could undermine the effectiveness of policies targeting specific populations or disadvantaged socioeconomic segments.

The linguistic framing analysis underscores the urgent need for language-specific calibration of LLMs. In an increasingly globalized and multilingual world, the development of models trained on linguistically and culturally diverse datasets is critical. Policymakers and researchers should approach the use of AI tools for behavioral analysis with caution, ensuring that the linguistic and cultural context of the target population is appropriately accounted for.

\subsection{Future Research Directions}

Our findings point to several promising directions for future research. First, this study examined risk attitude using the context-free approach, via lottery choice games. Future work should evaluate the predictive performance of LLMs in a broader range of decision-making scenarios under uncertainty, as well as other types of complex decision-making processes, to enable broader validation of these models. Second, exploring individual-level heterogeneity and implementing personalized calibration of LLM outputs based on respondents’ sociodemographic and psychological characteristics may improve the accuracy of simulated decision-making and enhance the practical utility of these models. In conclusion, this study represents an initial attempt to examine how LLMs simulate human behavior under uncertainty. While LLMs show promise as tools for behavioral modeling, there is still a considerable distance to go before they can fully and reliably replicate the complexity of human decision-making.

\section*{Data availability}

Due to legal and ethical considerations, the real-world case data are not publicly available; however, aggregated datasets can be obtained from the corresponding author upon request.

\section*{Code availability}

Due to confidentiality restrictions, the source code is not publicly available; however, access may be granted for legitimate research purposes upon contacting the corresponding author.

\section*{Author contributions}
\noindent
\textbf{Bing Song}: Conceptualization, Investigation, Methodology, Writing – original draft; 
\textbf{Jianing Liu}: Conceptualization, Investigation, Methodology, Writing – original draft; 
\textbf{Chenyang Wu}: Conceptualization, Investigation, Methodology, Fund acquisition, Writing – review \& editing; 
\textbf{Sisi Jian}: Conceptualization, Investigation, Methodology, Supervision, Fund acquisition, Writing – review \& editing; 
\textbf{Vinayak Dixit}: Data acquisition, Writing – review \& editing.

\section*{Competing interests}
The authors declare no competing interests

\bibliography{sn-bibliography}

\end{document}